\documentclass[table]{article} 
\usepackage{iclr2025_conference,times}
\usepackage{graphicx}
\usepackage{subcaption}
\usepackage{hyperref}
\usepackage{natbib}
\usepackage{float}
\usepackage{ulem}       
\usepackage{booktabs} 
\definecolor{Teal}{RGB}{80,139,243}
\hypersetup{
    colorlinks = true,
    citecolor = Teal,     
}

\usepackage{amsmath,amsfonts,bm}

\def\eqref#1{equation~\ref{#1}}

\def\1{\bm{1}}

\DeclareMathAlphabet{\mathsfit}{\encodingdefault}{\sfdefault}{m}{sl}
\SetMathAlphabet{\mathsfit}{bold}{\encodingdefault}{\sfdefault}{bx}{n}

\usepackage{hyperref}
\usepackage{url}

\title{Livatar-1: Real-Time Talking Heads Generation with Tailored Flow Matching}

\author{
\textbf{Haiyang Liu}\thanks{Equal Contribution} \quad 
\textbf{Xiaolin Hong}\textsuperscript{*} \quad
\textbf{Xuancheng Yang}\textsuperscript{*} \quad 
\textbf{Yudi Ruan}\textsuperscript{*} \quad \\
\hspace{0.02cm} \textbf{Xiang Lian} \quad 
\textbf{Michael Lingelbach} \quad
\textbf{Hongwei Yi} \quad
\textbf{Wei Li}\thanks{Project Lead}  \\
\hspace{0.02cm} Hedra Inc. \\
}

\iclrfinalcopy

\begin{document}
\maketitle

\begin{abstract}
We present \textbf{Livatar}, a real-time audio-driven talking heads videos generation framework. Existing baselines suffer from limited lip-sync accuracy and long-term pose drift. We address these limitations with a flow matching based framework. Coupled with system optimizations, \textbf{Livatar} achieves competitive lip-sync quality with a 8.50 LipSync Confidence on the HDTF dataset, and reaches a throughput of 141 FPS with an end-to-end latency of 0.17s on a single A10 GPU. This makes high-fidelity avatars accessible to broader applications. Our project is available at \url{https://www.hedra.com/} with with examples at \url{https://h-liu1997.github.io/Livatar-1/}.
\end{abstract}

\section{Introduction}
\label{sec.intro}

Recent breakthroughs in Large Language Models (LLMs)~\citep{touvron2023llama,openai2023gpt4,anil2023palm2} and real-time Text-to-Speech (TTS)~\citep{kim2021vits,ren2020fastspeech2,shen2018tacotron2} systems have paved the way for highly interactive, streaming AI agents. To truly unlock their potential, these AI agents require visual embodiments, enabling new applications in education, sales, and virtual companionship.

A typical scenario involves LLMs and TTS systems generating a streaming audio response based on a user's input. The remaining key problem to realizing these visualized agents is a real-time and streaming model that can generate talking-head videos from a single image and the streaming audio. 

Current approaches have two key problems: limited lip-sync accuracy and long-term pose drift, where cumulative errors cause the head's pose and shape to deviate over time. In this work, we present \textbf{Livatar}\footnote{This technical report is a shorten version only summarizing the performance of the Livatar due to the intellectual property policy, the complete version was finished earlier.}, a system designed to address these challenges and achieve production-ready quality and performance.

\section{Experiments}
\label{sec:experiments}

We focus on automated, no-reference metrics for evaluation. Following the evaluation protocol of recent video generation benchmarks~\citep{huang2024vbench}, we assess our method across four key dimensions: lip-sync quality~\citep{Chung16a_SyncNet}, content similarity~\citep{radford2021learning}, image quality~\citep{huang2024vbench}, and motion dynamics~\citep{huang2024vbench}.  

We compare \textbf{Livatar} with several leading talking-head generation models: SadTalker~\citep{zhang2023sadtalker}, Real3DPortrait~\citep{ye2024real3d}, 
Hallo3~\citep{cui2024hallo3}, Sonic~\citep{ji2025sonic}, and our reproduced INFP~\citep{zhu2025infp}. For evaluation, we construct a unified test set by randomly sampling 100 clips each from the HDTF and our internal datasets, these clips were filtered out from training. All input faces are cropped and resized to 512x512. 

As shown in Table~\ref{tab:compare_with_sota}, our method shows best lip-sync performance over all baselines. More video results are available on our project page.

\begin{table}[t]
    \centering
    \caption{\textbf{Comparison with existing methods.} 
    We compare our method with state-of-the-art video diffusion models (\textit{offline}) and other methods (\textit{realtime}) on the HDTF-100 (\textit{left}) and our Internal-100 (\textit{right}) test sets. $*$ denotes our reproduced version. CS and WR are Content Similarity and user study Win Rate (others vs. ours), respectively.}
    \vspace{0.2cm}
    \label{tab:compare_with_sota}
    \resizebox{0.999\linewidth}{!}{
    \begin{tabular}{l c cccc cccc c}
        & \textbf{Cost} & \textbf{Sync-C}$\uparrow$ & \textbf{CS}$\uparrow$ & \textbf{Quality}$\uparrow$ & \textbf{Dynamic}$\uparrow$
        & \multicolumn{1}{c}{\textbf{Sync-C}$\uparrow$} & \textbf{CS}$\uparrow$ & \textbf{Quality}$\uparrow$ & \textbf{Dynamic}$\uparrow$ & \multicolumn{1}{c}{\textbf{WR\%}} \\
        \midrule
        \textcolor{gray}{GroundTruth} 
        & & \textcolor{gray}{7.614} & \textcolor{gray}{0.928} & \textcolor{gray}{0.642} & \multicolumn{1}{c|}{\textcolor{gray}{0.660}}
        & \textcolor{gray}{6.995} & \textcolor{gray}{0.885} & \textcolor{gray}{0.656} & \textcolor{gray}{0.782} & \multicolumn{1}{|c}{-} \\
        Hallo3~\citep{cui2024hallo3}  
        & \textit{offline} & 6.814 & 0.915 & 0.638 & \multicolumn{1}{c|}{\textbf{0.870}}
        & 6.093 & 0.895 & 0.633 & \uline{0.792} & \multicolumn{1}{|c}{26.8} \\
        Sonic~\citep{ji2025sonic} 
        & \textit{offline} & \uline{8.495} & 0.935 & 0.626 & \multicolumn{1}{c|}{0.600}
        & \uline{7.998} & 0.916 & 0.616 & \textbf{0.832} & \multicolumn{1}{|c}{42.4} \\
        \midrule
        SadTalker~\citep{zhang2023sadtalker} 
        & \textit{realtime} & 6.704 & \textbf{0.965} & \textbf{0.697} & \multicolumn{1}{c|}{0.080}
        & 6.547 & \textbf{0.961} & \textbf{0.687} & 0.020 & \multicolumn{1}{|c}{4.5} \\
        Real3DPortrait~\citep{ye2024real3d} 
        & \textit{realtime} & 6.811 & 0.943 & 0.637 & \multicolumn{1}{c|}{0.030}
        & 6.529 & 0.934 & 0.602 & 0.000 & \multicolumn{1}{|c}{6.8} \\
        INFP*~\citep{zhu2025infp} 
        & \textit{realtime} & 7.357 & 0.928 & 0.633 & \multicolumn{1}{c|}{0.780}
        & 6.635 & 0.907 & 0.612 & 0.690 & \multicolumn{1}{|c}{28.9} \\
        \textbf{Livatar (Ours)}
        & \textit{realtime} & \textbf{8.501} & \uline{0.944} & \uline{0.645} & \multicolumn{1}{c|}{\uline{0.800}}
        & \textbf{8.014} & \uline{0.940} & \uline{0.636} & 0.772 & \multicolumn{1}{|c}{\textbf{50.0}} \\
    \end{tabular}}
\end{table}

\begin{figure*}[t]
\begin{center}
\includegraphics[width=1.0\textwidth]{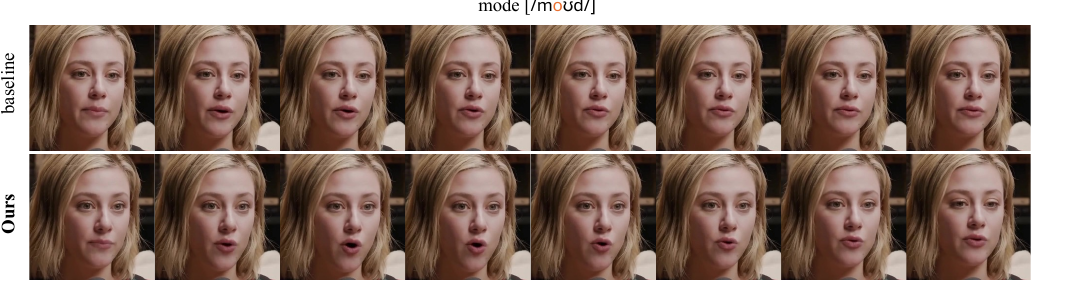}
\end{center}
\vspace{-0.3cm}
\caption{\textbf{Lip Synchronization Comparison.} Compare with baseline method, Livatar better handles mouth movements for sounds with strong lip closures, like plosives.
}
\vspace{-0.3cm}
\label{fig:exp_w}
\end{figure*}

\begin{figure}[H]
\begin{center}
\includegraphics[width=1.0\textwidth]{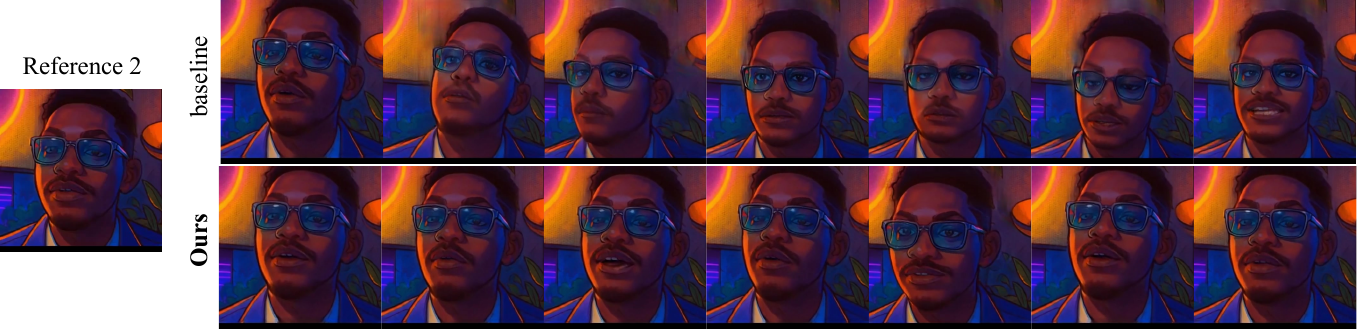}
\end{center}
\vspace{-0.3cm}
\caption{\textbf{Long Video Generation Comparison.} Compare with baseline method, Livatar generates long videos with improved appearance consistency.
}
\vspace{-0.3cm}
\label{fig:anchor}
\end{figure}

We implement several system-level optimizations to achieve real-time performance. These optimizations cumulatively reduce the inference latency for a single chunk (generating 24 new frames) from a baseline of 1.1s to 0.17s on an A10 GPU. This achieves a final throughput of 141 FPS. Compared with offline methods, which take around 20 seconds of end-to-end latency and achieve a 0.1 FPS throughput on an H100~\citep{kong2024hunyuanvideo}, our method is more efficient for real-time interaction. 

\section{Conclusion}
We presented \textbf{Livatar}, a system that generates real-time, streaming talking-head videos from a single image and an audio signal. Our system addresses two limitations of existing models, \textit{i.e.,} limited lip-sync accuracy and long-term pose drift. Our extensive experiments show that \textbf{Livatar} achieves competitive performance in both lip-sync quality and inference speed on consumer-grade hardware, demonstrating its suitability for practical, real-world applications.


\newpage
\bibliography{iclr2025_conference}
\bibliographystyle{iclr2025_conference}
\end{document}